\pdfoutput=1

\documentclass[11pt]{article}
\usepackage{tablefootnote}
\usepackage{hyperref}
\usepackage{authblk}  

\usepackage{listings}
\usepackage{xcolor}
\usepackage{float}
\usepackage{caption}

\usepackage{ACL2023}

\usepackage{times}
\usepackage{latexsym}
\usepackage{booktabs}
\usepackage{amsmath}

\usepackage[T1]{fontenc}

\usepackage[utf8]{inputenc}

\usepackage{microtype}

\usepackage{inconsolata}

\usepackage{graphicx}

\title{MRT at SemEval-2025 Task 8: Maximizing Recovery from Tables with Multiple Steps}

\author{
 \textbf{Maximiliano Hormazábal Lagos\textsuperscript{\textdagger}},
 \textbf{Álvaro Bueno Sáez\textsuperscript{\textdagger}},
 \textbf{Héctor Cerezo-Costas\textsuperscript{\textdagger}}, \\
 \textbf{Pedro Alonso Doval\textsuperscript{\textdagger}},
 \textbf{Jorge Alcalde Vesteiro\textsuperscript{\textdagger}} \\

  \small{
    \texttt{mhormazabal@gradiant.org}, \texttt{abueno@gradiant.org}, \texttt{hcerezo@gradiant.org}} \\
    \small{\texttt{palonso@gradiant.org}, \texttt{jalcalde@gradiant.org}}
  
}

\affil{\textsuperscript{\textdagger}Fundación Centro Tecnolóxico de Telecomunicacións de Galicia (GRADIANT), Vigo, Spain}

\begin{document}
\maketitle
\begin{abstract}
In this paper we expose our approach to solve the \textit{SemEval 2025 Task 8: Question-Answering over Tabular Data} challenge. Our strategy leverages Python code generation with LLMs to interact with the table and get the answer to the questions. The process is composed of multiple steps: understanding the content of the table, generating natural language instructions in the form of steps to follow in order to get the answer, translating these instructions to code, running it and handling potential errors or exceptions. These steps use open source LLMs and fine grained optimized prompts for each task (step). With this approach, we achieved a score of $70.50\%$ for subtask 1.

\end{abstract}


\section{Introduction}









Contemporary Natural Language Processing (NLP) is limited by the volume of information (text) that can be processed effectively while maintaining contextual relevance. During response generation this constraint impacts the recall of data needed to produce correct and complete answers \cite{liu-etal-2024-rethinking}. Tabular data exemplifies this challenge in particular, since it is the day-to-day task most affected by this restriction \cite{ruan2024languagemodelingtabulardata}. This paper addresses the SemEval 2025 Task 8: Question-Answering over Tabular Data \cite{osesgrijalba-etal-2025-semeval-2025}.

In this paper we present  Maximizing Recovery from Tables with Multiple Steps (MRT), a multi-step pipeline that leverages both LLMs and Python code generation to answer questions in the most factual way possible. Instead of an end-to-end strategy our system implements a sequential divide and conquer approach in which at every step either LLMs or heuristics are executed. Those steps cover from describing the tables (frequent values, column descriptions, statistical information), generating the list of instructions (in plain natural language) to carry the task and obtain the result, code execution and answer parsing.


We achieve $70.50\%$ accuracy in the Databench Challenge test set using this approach. The code that generated these results is publicly available\footnote{\url{https://github.com/Gradiant/MRT_TableQA/releases/tag/v1.0.0}}.

\section{Background}

Question answering (QA) focuses on retrieving accurate answers from \cite{wang2025accurateregretawarenumericalproblem} data sets. Recent methods for QA on tabular data, such as TAPAS \cite{Herzig_2020}, integrate transformers with architectures specifically tuned to extract answers directly from the tables used as context. However, LLMs have also been employed in zero-shot or few-shot strategies, since they are able to respond with a certain quality due to their prior knowledge and thus reduce the need for domain-specific fine-tuning of each domain \cite{kadam2020review}. Recent LLMs have demonstrated emergent reasoning capability, but still present difficulties with complex queries involving multiple columns, large tables, or ambiguous interpretations of a question.

Another approach is to parse natural language queries and transform them into formal queries such as SQL. Systems such as Seq2SQL \cite{zhong2017seq2sqlgeneratingstructuredqueries} or TableGPT2 \cite{yang2024evaluating} are designed to generate SQL queries from relational database queries or Python code, respectively. These methods offer advantages such as greater flexibility, as they are theoretically independent of the table size (which might not fit entirely in the context window of an LLM), and greater transparency, by including an intermediate step that allows auditing and reviewing the generated queries.

To evaluate these models, reference datasets have been critical. Wikipedia-based sets, such as WikiSQL \cite{zhong2017seq2sqlgeneratingstructuredqueries} and TabFact \cite{chen2020tabfactlargescaledatasettablebased}, provide structured evaluation environments but do not reflect the heterogeneity of real-world tabular data \cite{hwang2019comprehensiveexplorationwikisqltableaware}. In response, DataBench \cite{oses-grijalba-etal-2024-question} has been developed, which brings together $65$ real-world datasets with more than 1,300 manually crafted question-answer pairs across multiple domains.

Works such as TableRAG \cite{chen2024tableragmilliontokentableunderstanding} propose the use of RAG systems for tabular comprehension tasks, such as QA, employing techniques such as query expansion and a double transformation to query languages. This process translates, on the one hand, the schema to be interacted with and, on the other hand, the operation necessary to identify the cells with the answer. Also noteworthy are proposals such as Chain-of-Table \cite{wang2024chainoftableevolvingtablesreasoning}, which implements Chain-of-Thought as an iterative reasoning mechanism. Instead of executing code in one shot, operations are executed in each iteration to add or discard information from the table until the answer is found.

Despite recent progress, some challenges still remain, such as improving reasoning within multiple rows, handling different domains and languages or integrating several tables and increasing the explainability of the full process.

\section{System Overview}

 The strategy developed in our system consists of loading the table as a Pandas dataframe, and then, with the use of LLMs, generating Python code to interact with the tabular data to finally obtain the answer to each question. 
 For this, we implemented multiple modules that are executed sequentially for each question. Some of these steps are heuristics, whereas others are LLM-based.
 \\ \\
 Each of the modules can use different LLMs. For this work, we have used multilingual models from the families Qwen2.5 \cite{qwen2.5}, Qwen2.5-coder \cite{hui2024qwen2}, Llama-3 \cite{llama3modelcard} and Phi-4 \cite{abdin2024phi} among others.
 
\begin{figure}[h]
    \centering
    \includegraphics[width=0.5\textwidth]{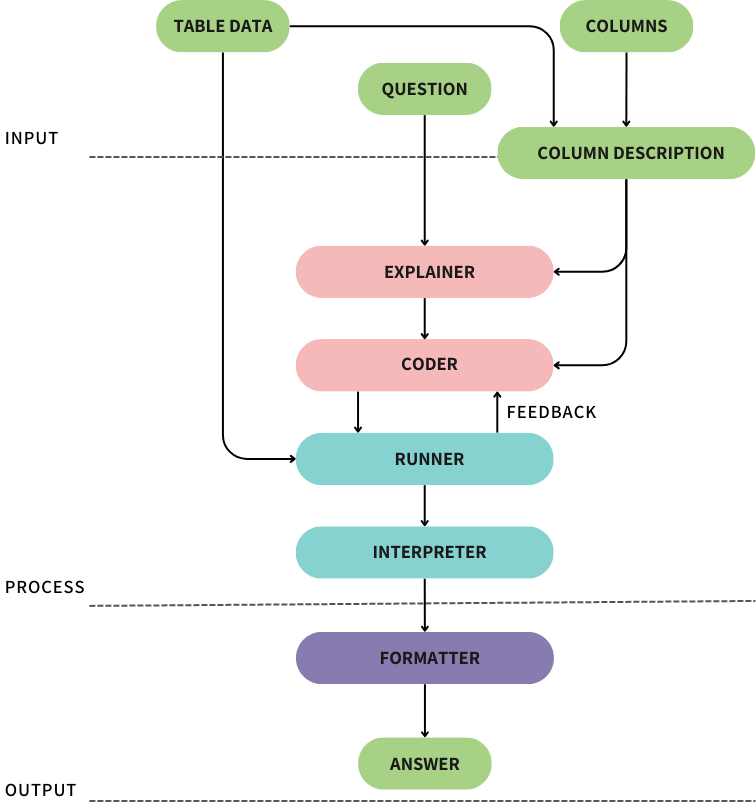}
    \caption{Diagram of the system showing all the steps involved in the generation of the response.}
    \label{fig:architecture}
\end{figure}
  
The Figure \ref{fig:architecture} depicts an overview of the workflow of the system. The first step is understanding and analyzing the information of the table (type of data, appearance of null values, etc.). Then, an LLM generates textual instructions with the reasoning steps to follow in order to get the answer. After that, a code generation model converts the instructions to Python code. Then, the Runner executes this code. If an exception occurs during code execution or answer parsing, the system steps back into the Coder in an iterative looping process until it gets a valid answer or a limit is exceeded. Finally, there are formatting steps that implement functions such as getting the answer in the desirable data type or selecting the correct number of decimals in numbers.


\subsection{Column Descriptor}
This module aims to analyze and understand the content of the table.
First, it analyzes the input table obtaining some statistical data for each column, such as the data type, the number of unique values, if it has missing values, the \textit{max}, \textit{min}, \textit{mean} values and \textit{standard deviation} (when it applies), and the most frequent values.

The second step involves serializing a subset of the table and prompting an LLM to describe the content of each column. While column names are usually descriptive, they can sometimes lack uniqueness, contain abbreviations, or be better explained within the context of the other columns.

The results of this module for each table are cached and hence this step is skipped for the following questions related to the same table. 

Examples of the output of this module are shown in Listing \ref{listing:column_description_Example} in Appendix I.

\subsection{Explainer}
 The Explainer module prompts an LLM to break down the steps required to answer a question using the table information. These instructions must be written in natural language. The prompt includes relevant details extracted by the Column Descriptor, such as column's name, description, value type, and whether it has missing values. For numeric data types, it includes their range, and for categorical types, it lists their values if there are fewer than a configured number of unique options (fixed to $7$), or otherwise just the most frequent values

The range of possible values is relevant for many questions that involve filtering by specific conditions. For example, to filter rows referring to a woman, the 'Gender' column might have various entries indistinctly like \textit{woman}, \textit{W}, \textit{female}, \textit{F}, etc. The same variety is observed in boolean values.

Guidelines are included in the prompt to force the system to use the exact given name of the columns, avoid the use of enumerations or to omit writing any code example.

The Explainer module includes a second step that prompts the LLM to review and refine the generated instructions. This process can help eliminate unnecessary steps or simplify them for greater precision.

Finally, the module parses the response to produce a list of strings, each representing an individual instruction.

Listing \ref{listing:explainer_example} in Appendix I contains examples of the explainer output.

\subsection{Coder and Runner}
The Coder module uses an LLM to generate Python code using the Pandas package that implements the natural language instructions in a method with the following header:

\begin{verbatim}
def parse_dataframe(df: pd.DataFrame) \
-> str:
    ...
\end{verbatim}

The prompt includes guidelines to avoid exceptions, such as using the exact column names, casting specific data types, generating a single Python method, and avoiding the Pandas \textit{groupby} function. The latter directive is based on empirical observations that the models we used tended to overuse this instruction, frequently resulting in numerous errors during code execution.

The LLM response is processed by a parser that employs heuristics to verify and standardize the various syntax generated by the model. As heuristics we employ different already preexisting libraries such as autopep8, autoflake, and lib\_23 to fix minimal inconsistencies in the Python code syntax. In particular lib\_23 is used to parse python 2 code into python 3, and will check for missing commas/paretheses (for example). We also employ the AST tree parsing to detect when something doesn't have python code format. When detected (via parsing or exception), the system makes up to four attempts to correct them by returning the response to the LLM for revision.

Finally, the Runner module executes the code generated by the Coder and returns the result. If an exception occurs during execution, the process reverts to the Coder, with a maximum of three retries, to regenerate the code. The exception is added to the prompt to prevent it in subsequent iterations.

\subsection{Interpreter}
The Interpreter module checks if the format of the answer matches the expected type of data for the question.
To achieve this, it first consults an LLM to determine the most suitable type of data to answer the question given the accepted types of the task: \textit{Boolean}, \textit{String}, \textit{Number}, \textit{List of Strings}, and \textit{List of Numbers}.

Then, in a second call to the LLM, asks it to fit the answer to the given format if it is not already correct. With this, we correct many errors like returning numbers of booleans casted to strings (see examples in Table \ref{table:interpreter_examples} in Appendix I).

\subsection{Formatter}
The last module in the workflow is the formatter. This module, based on rules is in charge of setting the answer in the most suitable format to match the expected task output. (see examples in Table \ref{table:formatter_examples} in Appendix I). For example, checks the data type of the answer, and casts it or make some format transformations. 

In most cases, this module does not require any modifications due to the correction performed in previous steps. However, in certain instances, adjustments are necessary to ensure alignment with the gold labels during task evaluation.



	
	
	
	
	

\section{Experimental Setup}


The experimental setup consisted of two approaches for executing the modules plus the combination of configurations of each of the modules. Initially, the modules were run in series, as lighter models that could fit concurrently in memory were used. In this approach, each question for each table was processed sequentially through all system modules. However, during testing phase with heavier models, it became necessary to implement a system that allowed to load and unload the models as it was required by each step. This approach executes each of the step in batches. All the questions are processed for each step before passing to the next one. Hence only the model used in each step is loaded in memory. This allows to load bigger models for each of the steps whilst using the same GPU without involving excessive overhead in loading/unloading models.

The hardware used to run the tests was an NVIDIA RTX-a6000 that combines 84 second-generation RT cores, 336 third-generation Tensor cores, and 10,752 CUDA cores with 48 GB of graphics memory for performance.  


In  \ref{subsec:models}, we define the model configurations used in the test phase. During development, we also used reduced versions of these models with $8B$ parameters.

\subsection{Dataset splits}

Although no training of any model has been performed, the splits of the dataset are shown below. 

\begin{table}[h!]
\centering
\begin{tabular}{|c|c|c|}
\hline
\textbf{Split} & \textbf{Tables} & \textbf{Questions} \\
\hline
train & 49 & 988 \\
\hline
dev & 16 & 320 \\
\hline
test & 15 & 522 \\
\hline

\end{tabular}
\caption{Distribution of number of tables and questions for each split in the dataset}
\label{table:dataset_splits}
\end{table}

Train, dev and test splits have been used for the development of the modules. 

\subsection{Models} \label{subsec:models}

Llama 3\footnote{https://huggingface.co/meta-llama}, Phi\footnote{https://huggingface.co/microsoft} and Qwen\footnote{https://huggingface.co/Qwen} models of different sizes have been used for the different modules of the system. 

For Llama 3 only the 8B size model was used\footnote{https://huggingface.co/meta-llama/Meta-Llama-3-8B-Instruct}. For Phi-4, the 14B version was choosen\footnote{https://huggingface.co/microsoft/phi-4} and, finally, the main family of models used in the tests is Qwen. Two types of Qwen models have been executed: Qwen25 and Qwen25 code. 

For Qwen25 two sizes have been selected: 7B\footnote{https://huggingface.co/Qwen/Qwen2.5-7B-Instruct} and 14B\footnote{https://huggingface.co/Qwen/Qwen2.5-14B-Instruct}. For Qwen25 code also the same two sizes were used: 7B\footnote{https://huggingface.co/Qwen/Qwen2.5-Coder-7B-Instruct} and 14B\footnote{https://huggingface.co/Qwen/Qwen2.5-Coder-14B-Instruct}. 

    

Let us emphasize that the $8B$ models have been used mainly in the first battery of tests and development whilst the $14B$ models were used in the final execution of the system.

\subsection{Test and configuration}

Table \ref{table:experiments} summarizes the different experiments performed, with the model used in each module for each experiment. 

\begin{table}[h!]
\centering
\resizebox{\columnwidth}{!}{
\begin{tabular}{|c|c|c|}
\hline
 \textbf{Explainer} & \textbf{Coder} & \textbf{Interpreter} \\
\hline
$llama3_{8B}$ & $qwen25_{14B}$\_code & $qwen25_{14B}$ \\
\hline
$phi4_{14B}$ & $qwen25_{14B}$\_code & $qwen25_{14B}$ \\
\hline
$qwen25_{14B}$ & $qwen25_{14B}\_code$ & $qwen25_{14B}$ \\
 \hline  

\end{tabular}
}
\caption{Different configurations for each step that uses LLMs}
\label{table:experiments}
\end{table}

 Table \ref{table:experiments} shows that, for the most part, experiments have been performed keeping the \textit{Qwen25 14B-coder} model in the Coder module and selecting different models for the Explainer module. The \textit{Llama}, \textit{Phi-4}, \textit{Qwen2.5 14B}, and \textit{Qwen2.5 7B} models have been tested in the Explainer, whereas \textit{Qwen2.5 7B} was selected to be the main interpreter mainly because during the development time small tests were performed in that module with both Llama and \textit{Phi-4} and the results were not remarkable. Finally, note that the Colum Descriptor module is not in the table to save space. The module used for Column Descriptor was \textit{Qwen2.5 7B} in all experiments and was run separately because the column description process is done per table and not per question. 

	
	
	
	
	

\section{Results}

\subsection{Performance in Validation split}

Table \ref{table:acc_per_step} shows the accuracy of the system using different models. The model indicates the one used in the Explainer. For the Column Descriptor and Coder all the models share the \textit{Qwen2.5 7B} and \textit{Qwen2.5 14B-Coder} respectively. The Explainer step is performed by the \textit{Qwen2.5 14B}. The Ensemble is obtained by the majority voting of the three models. Ties are resolved prioritizing \textit{Qwen2.5} answers over \textit{Phi-4} and \textit{Llama} models.

\begin{table}[h!]
\centering
\begin{tabular}{|c|c|c|c|}
\hline
\textbf{Models} & \textbf{Run} & \textbf{Interpret} &
\textbf{Format} \\
\hline
$\text{llama3}_{8B}$ & 0.563 & 0.613 & 0.606 \\
\hline
$\text{phi4}_{14B}$ & 0.756 & 0.756 & 0.75 \\
\hline
$\text{qwen25}_{14B}$ & 0.762 & 0.766 & 0.759 \\
\hline
$\text{Ensemble}_{\text{max}}$ & 0.778 & 0.772 & 0.766 \\
\hline
\end{tabular}
\caption{Accuracy of the different strategies in the development split using the outputs of the Runner, Interpreter and Formatter step.}
\label{table:acc_per_step}
\end{table}

As can be seen in the results, the heuristics to format the final predictions sometimes introduce additional errors.

If we filter the prediction by type of response requested (Table \ref{table:acc_per_type_dev}) we can see that the lists of items (either number or categorical) have a much higher difficulty than the singular responses. As expected boolean responses are easy to handle by the system as only two values are possible. Nevertheless the best individual model, \textit{Qwen2.5 14B}, obtained better results in the categorical answers. 

\begin{table}[ht]
\centering
\begin{tabular}{|c|c|c|c|c|}
\hline
\textbf{Answer} & \textbf{llama3} & \textbf{phi4} &
\textbf{qwen25} & \textbf{Ens} \\
\textbf{Type} & \bf 8B & \bf 14B & \bf 14B & \bf emble \\
\hline
Boolean & 0.75 & 0.844 & 0.828 & \bf 0.844 \\
\hline
Number & 0.627 & \bf 0.851 & 0.761 & 0.836 \\
\hline
 Categ. & 0.639 & 0.754 & \bf 0.836 & 0.787 \\
\hline
 $\text{List}_{\text{Num}}$ & 0.538 & 0.692 & 0.754 & \bf 0.769 \\
\hline
 $\text{List}_{\text{Cat}}$ & 0.476 & 0.603 & \bf 0.619 & 0.587 \\
\hline
All & 0.606 & 0.75 & 0.759 & \bf 0.766 \\
\hline
\end{tabular}
\caption{Accuracy of the different strategies per data type of the expected answer in the validation split}
\label{table:acc_per_type_dev}
\end{table}

After performing the same analysis in the test split (\ref{table:acc_per_type_test}), we can see that in general all the models experience a poorer performance in all categories but the ranking are almost the same. 

\begin{table}[ht]
\centering
\begin{tabular}{|c|c|c|c|c|}
\hline
\textbf{Answer} & \textbf{llama3} & \textbf{phi4} &
\textbf{qwen2.5} & \textbf{Ens} \\
\textbf{Type} & \bf 8B & \bf 14B & \bf 14b & \bf emble \\
\hline
Boolean & 0.659 & 0.791 & \bf 0.829 &  0.814 \\
\hline
Number & 0.417 & \bf 0.628 & 0.596 & 0.596 \\
\hline
 Categ. & 0.459 & 0.635 &  0.703 & \bf 0.716 \\
\hline
 $\text{List}_{\text{Num}}$ & 0.407 & 0.560 & 0.615 &  \bf 0.626 \\
\hline
 $\text{List}_{\text{Cat}}$ & 0.417 & 0.514 &  0.542 & \bf 0.556 \\
\hline
All & 0.480 & 0.642 & 0.665 & \bf 0.667 \\
\hline
\end{tabular}
\caption{Accuracy of the different strategies per data type of the expected answer in the test split}
\label{table:acc_per_type_test}
\end{table}

\subsection{Manual Error Analysis}

We performed a manual error analysis of the results flagged as an error by the official evaluator for the Qwen2.5. The results are summarized in Table \ref{table:manual_error_analysis}. 

When passing from instructions to code some of them are usually omitted or not treated properly. That was always true when the user requested to resolve ties in alphabetical order. Certain operations such as "group-by" were avoided in the prompt as the Coder was less capable of consistently generating error-free code when using them. Although our solution uses the prompt to discourage the LLM to use certain functions, another interesting option is to provide alternative implementations for them.

One clear flaw of the current design is that the system fails to filter by certain values when they do not appear in the common values of the column (e.g. when asked for \textit{Biden} the system does not know that this value appears as \textit{Joe Biden} in the table unless it is in the frequent values given by the Column Descriptor. This accounts for $15\%$ of the errors. This type of errors could be mitigated by linking values asked in the query with the real ones appearing in the table during a pre-processing step (e.g. after the natural language instructions are given). Formatting issues of the response are almost $10\%$ of the remaining errors. Handling these errors requires careful implementation of additional post-processing heuristics, and it relates greatly to how the metric to measure the performance is actually implemented (e.g. rounding issues, partial match, how lists are expected, if ordering of the elements is taking into account or not, etc.).

The \textit{others} set accounts for all unclassifiable errors, in general due to ambiguous questions or expected answers or incorrect ground truth samples.

\begin{table}[h!]
\centering
\begin{tabular}{|c|c|}
\hline
\textbf{Description} & \textbf{\% error} \\
\hline
Wrong cell value filtering & 14.29 \\ 
\hline
Wrong Instructions & 37.66 \\
\hline
Wrong code (incl. exceptions) & 14.29 \\
\hline
Formatting (transformations) & 6.49 \\
\hline
Formatting (answer type) & 3.90 \\
\hline
Others & 23.38 \\
\hline
\end{tabular}
\caption{Manual analysis of the errors of the Qwen14 model in the validation split}
\label{table:manual_error_analysis}
\end{table}

\section{Conclusion}

This paper has addressed our proposal, MRT, in response to the challenge proposed in Semeval 2025 regarding tabular data question-answering. This technique, which introduced a multi-step pipeline leveraging both LLMs and their ability for code generation, achieved a $70.50\%$ accuracy. 

Despite the competitive results, MRT is limited due to several factors, such as formatting the output correctly and some semantic ambiguities that are not interpreted correctly (e.g., double negation in the question). Nevertheless, the largest set of errors is due to an incorrect filter of the column values (either because the value type is incorrectly detected or because the value does not appear in the same way in the question and table). Additionally, MRT encounters difficulties in addressing abstract, more subjective, and less clear questions, which can be attributed to the size of the models employed.

Future work will focus on enhancing several of the modules to eliminate all accuracy losses introduced in the between-pipeline steps and adopting a less code-driven and more linguistic approach to ambiguous questions over tabular data.





\bibliography{custom}

\begin{thebibliography}{17}
\expandafter\ifx\csname natexlab\endcsname\relax\def\natexlab#1{#1}\fi

\bibitem[{Abdin et~al.(2024)Abdin, Aneja, Behl, Bubeck, Eldan, Gunasekar, Harrison, Hewett, Javaheripi, Kauffmann et~al.}]{abdin2024phi}
Marah Abdin, Jyoti Aneja, Harkirat Behl, S{\'e}bastien Bubeck, Ronen Eldan, Suriya Gunasekar, Michael Harrison, Russell~J Hewett, Mojan Javaheripi, Piero Kauffmann, et~al. 2024.
\newblock Phi-4 technical report.
\newblock \emph{arXiv preprint arXiv:2412.08905}.

\bibitem[{AI@Meta(2024)}]{llama3modelcard}
AI@Meta. 2024.
\newblock \href {https://github.com/meta-llama/llama3/blob/main/MODEL_CARD.md} {Llama 3 model card}.

\bibitem[{Chen et~al.(2024)Chen, Miculicich, Eisenschlos, Wang, Wang, Chen, Fujii, Lin, Lee, and Pfister}]{chen2024tableragmilliontokentableunderstanding}
Si-An Chen, Lesly Miculicich, Julian~Martin Eisenschlos, Zifeng Wang, Zilong Wang, Yanfei Chen, Yasuhisa Fujii, Hsuan-Tien Lin, Chen-Yu Lee, and Tomas Pfister. 2024.
\newblock \href {http://arxiv.org/abs/2410.04739} {Tablerag: Million-token table understanding with language models}.

\bibitem[{Chen et~al.(2020)Chen, Wang, Chen, Zhang, Wang, Li, Zhou, and Wang}]{chen2020tabfactlargescaledatasettablebased}
Wenhu Chen, Hongmin Wang, Jianshu Chen, Yunkai Zhang, Hong Wang, Shiyang Li, Xiyou Zhou, and William~Yang Wang. 2020.
\newblock \href {http://arxiv.org/abs/1909.02164} {Tabfact: A large-scale dataset for table-based fact verification}.

\bibitem[{Herzig et~al.(2020)Herzig, Nowak, Müller, Piccinno, and Eisenschlos}]{Herzig_2020}
Jonathan Herzig, Pawel~Krzysztof Nowak, Thomas Müller, Francesco Piccinno, and Julian Eisenschlos. 2020.
\newblock \href {https://doi.org/10.18653/v1/2020.acl-main.398} {Tapas: Weakly supervised table parsing via pre-training}.
\newblock In \emph{Proceedings of the 58th Annual Meeting of the Association for Computational Linguistics}. Association for Computational Linguistics.

\bibitem[{Hui et~al.(2024)Hui, Yang, Cui, Yang, Liu, Zhang, Liu, Zhang, Yu, Dang et~al.}]{hui2024qwen2}
Binyuan Hui, Jian Yang, Zeyu Cui, Jiaxi Yang, Dayiheng Liu, Lei Zhang, Tianyu Liu, Jiajun Zhang, Bowen Yu, Kai Dang, et~al. 2024.
\newblock Qwen2. 5-coder technical report.
\newblock \emph{arXiv preprint arXiv:2409.12186}.

\bibitem[{Hwang et~al.(2019)Hwang, Yim, Park, and Seo}]{hwang2019comprehensiveexplorationwikisqltableaware}
Wonseok Hwang, Jinyeong Yim, Seunghyun Park, and Minjoon Seo. 2019.
\newblock \href {http://arxiv.org/abs/1902.01069} {A comprehensive exploration on wikisql with table-aware word contextualization}.

\bibitem[{Kadam and Vaidya(2020)}]{kadam2020review}
Suvarna Kadam and Vinay Vaidya. 2020.
\newblock Review and analysis of zero, one and few shot learning approaches.
\newblock In \emph{Intelligent Systems Design and Applications: 18th International Conference on Intelligent Systems Design and Applications (ISDA 2018) held in Vellore, India, December 6-8, 2018, Volume 1}, pages 100--112. Springer.

\bibitem[{Liu et~al.(2024)Liu, Wang, and Chen}]{liu-etal-2024-rethinking}
Tianyang Liu, Fei Wang, and Muhao Chen. 2024.
\newblock \href {https://doi.org/10.18653/v1/2024.naacl-long.26} {Rethinking tabular data understanding with large language models}.
\newblock In \emph{Proceedings of the 2024 Conference of the North American Chapter of the Association for Computational Linguistics: Human Language Technologies (Volume 1: Long Papers)}, pages 450--482, Mexico City, Mexico. Association for Computational Linguistics.

\bibitem[{Os{\'e}s~Grijalba et~al.(2025)Os{\'e}s~Grijalba, Ure{~n}a-L{\'o}pez, Mart{'i}nez~C{\'a}mara, and Camacho-Collados}]{osesgrijalba-etal-2025-semeval-2025}
Jorge Os{\'e}s~Grijalba, Luis~Alfonso Ure{~n}a-L{\'o}pez, Eugenio Mart{'i}nez~C{\'a}mara, and Jose Camacho-Collados. 2025.
\newblock {S}em{E}val-2025 task 8: Question answering over tabular data.
\newblock In \emph{Proceedings of the 19th International Workshop on Semantic Evaluation (SemEval-2025)}, Vienna, Austria. Association for Computational Linguistics.

\bibitem[{Os{\'e}s~Grijalba et~al.(2024)Os{\'e}s~Grijalba, Ure{\~n}a-L{\'o}pez, Mart{\'i}nez~C{\'a}mara, and Camacho-Collados}]{oses-grijalba-etal-2024-question}
Jorge Os{\'e}s~Grijalba, L.~Alfonso Ure{\~n}a-L{\'o}pez, Eugenio Mart{\'i}nez~C{\'a}mara, and Jose Camacho-Collados. 2024.
\newblock \href {https://aclanthology.org/2024.lrec-main.1179/} {Question answering over tabular data with {D}ata{B}ench: A large-scale empirical evaluation of {LLM}s}.
\newblock In \emph{Proceedings of the 2024 Joint International Conference on Computational Linguistics, Language Resources and Evaluation (LREC-COLING 2024)}, pages 13471--13488, Torino, Italia. ELRA and ICCL.

\bibitem[{Ruan et~al.(2024)Ruan, Lan, Ma, Dong, He, and Feng}]{ruan2024languagemodelingtabulardata}
Yucheng Ruan, Xiang Lan, Jingying Ma, Yizhi Dong, Kai He, and Mengling Feng. 2024.
\newblock \href {http://arxiv.org/abs/2408.10548} {Language modeling on tabular data: A survey of foundations, techniques and evolution}.

\bibitem[{Wang et~al.(2025)Wang, Qi, and Gan}]{wang2025accurateregretawarenumericalproblem}
Yuxiang Wang, Jianzhong Qi, and Junhao Gan. 2025.
\newblock \href {http://arxiv.org/abs/2410.12846} {Accurate and regret-aware numerical problem solver for tabular question answering}.

\bibitem[{Wang et~al.(2024)Wang, Zhang, Li, Eisenschlos, Perot, Wang, Miculicich, Fujii, Shang, Lee, and Pfister}]{wang2024chainoftableevolvingtablesreasoning}
Zilong Wang, Hao Zhang, Chun-Liang Li, Julian~Martin Eisenschlos, Vincent Perot, Zifeng Wang, Lesly Miculicich, Yasuhisa Fujii, Jingbo Shang, Chen-Yu Lee, and Tomas Pfister. 2024.
\newblock \href {http://arxiv.org/abs/2401.04398} {Chain-of-table: Evolving tables in the reasoning chain for table understanding}.

\bibitem[{Yang et~al.(2024{\natexlab{a}})Yang, Yang, Zhang, Hui, Zheng, Yu, Li, Liu, Huang, Wei, Lin, Yang, Tu, Zhang, Yang, Yang, Zhou, Lin, Dang, Lu, Bao, Yang, Yu, Li, Xue, Zhang, Zhu, Men, Lin, Li, Xia, Ren, Ren, Fan, Su, Zhang, Wan, Liu, Cui, Zhang, and Qiu}]{qwen2.5}
An~Yang, Baosong Yang, Beichen Zhang, Binyuan Hui, Bo~Zheng, Bowen Yu, Chengyuan Li, Dayiheng Liu, Fei Huang, Haoran Wei, Huan Lin, Jian Yang, Jianhong Tu, Jianwei Zhang, Jianxin Yang, Jiaxi Yang, Jingren Zhou, Junyang Lin, Kai Dang, Keming Lu, Keqin Bao, Kexin Yang, Le~Yu, Mei Li, Mingfeng Xue, Pei Zhang, Qin Zhu, Rui Men, Runji Lin, Tianhao Li, Tingyu Xia, Xingzhang Ren, Xuancheng Ren, Yang Fan, Yang Su, Yichang Zhang, Yu~Wan, Yuqiong Liu, Zeyu Cui, Zhenru Zhang, and Zihan Qiu. 2024{\natexlab{a}}.
\newblock Qwen2.5 technical report.
\newblock \emph{arXiv preprint arXiv:2412.15115}.

\bibitem[{Yang et~al.(2024{\natexlab{b}})Yang, Yang, Jin, Miao, Zhang, Yang, Cui, Zhang, Hui, and Lin}]{yang2024evaluating}
Jian Yang, Jiaxi Yang, Ke~Jin, Yibo Miao, Lei Zhang, Liqun Yang, Zeyu Cui, Yichang Zhang, Binyuan Hui, and Junyang Lin. 2024{\natexlab{b}}.
\newblock Evaluating and aligning codellms on human preference.
\newblock \emph{arXiv preprint arXiv:2412.05210}.

\bibitem[{Zhong et~al.(2017)Zhong, Xiong, and Socher}]{zhong2017seq2sqlgeneratingstructuredqueries}
Victor Zhong, Caiming Xiong, and Richard Socher. 2017.
\newblock \href {http://arxiv.org/abs/1709.00103} {Seq2sql: Generating structured queries from natural language using reinforcement learning}.

\end{thebibliography}
\bibliographystyle{acl_natbib}


\appendix
\section{Appendix I: Examples of submodule outputs}
Examples of the outputs of the Column descriptor  (Listing \ref{listing:column_description_Example}) and Explainer (Listing \ref{listing:explainer_example}), and examples of transformations made in the Interpreter (Table \ref{table:interpreter_examples}) and Formatter (Table \ref{table:formatter_examples}) steps.

\lstdefinelanguage{JSON}{
    string=[s]{"}{"},
    stringstyle=\color{purple},
    numbers=left,
    numberstyle=\tiny,
    stepnumber=1,
    numbersep=8pt,
    showstringspaces=false,
    breaklines=true,
    frame=single,
    basicstyle=\ttfamily\small,
    keywordstyle=\color{blue},
    commentstyle=\color{green!60!black},
    literate=
     *{0}{{{\color{blue}0}}}{1}
      {1}{{{\color{blue}1}}}{1}
      {2}{{{\color{blue}2}}}{1}
      {3}{{{\color{blue}3}}}{1}
      {4}{{{\color{blue}4}}}{1}
      {5}{{{\color{blue}5}}}{1}
      {6}{{{\color{blue}6}}}{1}
      {7}{{{\color{blue}7}}}{1}
      {8}{{{\color{blue}8}}}{1}
      {9}{{{\color{blue}9}}}{1}
      {:}{{{\color{black}{:}}}}{1}
      {,}{{{\color{black}{,}}}}{1}
      {\{}{{{\color{black}{\{}}}}{1}
      {\}}{{{\color{black}{\}}}}}{1}
      {[}{{{\color{black}{[}}}}{1}
      {]}{{{\color{black}{]}}}}{1},
}


\begin{figure}[h!]
\begin{lstlisting}[language=JSON]
{
"name": "trip_distance",
"type": "float64",
"missing_values": 0,
"unique": 1259,
"flag_binary": false,
"mean": 2.0519498,
"std": 1.6832561884020858,
"freq_values": null,
"description": {
  "name": "trip_distance",
  "description": "Distance of the taxi
  trip, typically measured in miles or
  kilometers."
 }
},

{
"name": "Have you ever use an online 
         dating app?",
"type": "category",
"missing_values": 0,
"unique": 2,
"flag_binary": false,
"mean": 0.0,
"std": 0.0,
"freq_values": [
    "Yes",
    "No"
],
"description": {
  "name": "Have you ever use an online 
  dating app?",
  "description": "Indicates whether 
  the respondent has ever used an 
  online dating application."
 }
}
\end{lstlisting}
\caption{Examples of outputs of the Column Descriptions for two columns.} 
\label{listing:column_description_Example}
\end{figure}

\begin{figure}[h!]
\begin{lstlisting}[language=JSON]
Question: "What is the primary type of
the Pok\'{e}mon with the highest defense 
stat?"

Explainer Output:
['Sort the rows in descending order 
based on the "defense" column', 
 'Select the row at the top of the 
sorted list', 
 'Access the "type1" column of 
the selected row', 
 'Return the value in the "type1"
column as the answer.']
\end{lstlisting}
\caption{Examples of outputs of the explainer} 
\label{listing:explainer_example}
\end{figure}

\begin{table}[H]
\centering
\resizebox{\columnwidth}{!}{
\begin{tabular}{|c|c|c|}
\hline
\textbf{Input} & \textbf{Expected} & \textbf{Output}  \\
\hline
"False" & Boolean & False \\ 
\hline
True & Boolean & True (unchanged) \\ 
\hline
1, 21, 14 & List of numbers & \texttt{[}1, 21, 14\texttt{]} \\
\hline
Water, Normal  & List of strings & \texttt{[}"Water", "Normal"\texttt{]} \\
\hline
\texttt{[}"16.0", "1.0"\texttt{]} & List of numbers & \texttt{[}16.0, 1.0\texttt{]} \\
\hline
0.2748 & Number & 0.2748 (unchanged) \\
\hline
\end{tabular}
}
\caption{Examples of transformations in the Interpreter}
\label{table:interpreter_examples}
\end{table}

\begin{table}[H]
\centering
\resizebox{\columnwidth}{!}{
\begin{tabular}{|c|c|}
\hline
\textbf{Input} & \textbf{Output}  \\
\hline
2.0  & 2 \\ 
\hline
\texttt{[}38.0, 23.0, 39.0\texttt{]} & \texttt{[}38, 23, 39\texttt{]}\\
\hline
(1000, 2000, 3000) & \texttt{[}1000, 2000, 3000\texttt{]} \\
\hline
400 & 400 (no changes) \\
\hline
\end{tabular}
}
\caption{Examples of transformations in the Formatter}
\label{table:formatter_examples}
\end{table}



\end{document}